\documentclass[10pt,english,review, 3p]{elsarticle}
\usepackage[T1]{fontenc}
\UseRawInputEncoding
\usepackage{color}
\usepackage{babel}
\usepackage{array}
\usepackage{float}
\usepackage{multirow}
\usepackage{algorithm2e}
\usepackage{amsmath}
\usepackage{amsthm}
\usepackage{amssymb}
\usepackage{graphicx}
\usepackage[bookmarks=false,
 breaklinks=false,pdfborder={0 0 1},backref=false,colorlinks=true]
 {hyperref}
\hypersetup{
 linkcolor=blue, citecolor=blue, urlcolor=blue}

\makeatletter

\providecommand{\tabularnewline}{\\}

\usepackage{times}
\usepackage{hyperref}
\usepackage[symbol]{footmisc}

\ifdefined\showcaptionsetup
 \PassOptionsToPackage{caption=false}{subfig}
\fi
\usepackage{subfig}
\makeatother

\begin{document}
\begin{frontmatter}
\title{Predicting the Reliability of an Image Classifier under Image Distortion}
\author{Dang Nguyen\thanks{Corresponding author}, Sunil Gupta, Kien Do, Svetha
Venkatesh}
\address{Applied Artificial Intelligence Institute (A\textsuperscript{2}I\textsuperscript{2}),
Deakin University, Geelong, Australia\\
\{d.nguyen, sunil.gupta, k.do, svetha.venkatesh\}@deakin.edu.au}
\begin{abstract}
In image classification tasks, deep learning models are vulnerable
to image distortions i.e. their accuracy significantly drops if the
input images are distorted. An\textit{ image-classifier} is considered
``\textit{reliable}'' if its accuracy on distorted images is above
a user-specified threshold. For a quality control purpose, it is important
to predict if the image-classifier is unreliable/reliable under a
distortion level. In other words, we want to predict whether a distortion
level makes the image-classifier ``non-reliable'' or ``reliable''.
Our solution is to construct a training set consisting of distortion
levels along with their ``non-reliable'' or ``reliable'' labels,
and train a machine learning predictive model (called\textit{ distortion-classifier})
to classify unseen distortion levels. However, learning an effective
distortion-classifier is a challenging problem as the training set
is \textit{highly imbalanced}. To address this problem, we propose
a Gaussian process based method to rebalance the training set. We
conduct extensive experiments to show that our method significantly
outperforms several baselines on six popular image datasets.
\end{abstract}
\begin{keyword}
Image classification; Reliability prediction; Image distortion; Imbalance
classification; Gaussian process.
\end{keyword}
\end{frontmatter}
\global\long\def\transp#1{\transpose{#1}}%
 
\global\long\def\transpose#1{#1^{\mathsf{T}}}%
\global\long\def\argmax#1{\underset{#1}{\text{argmax}\ } }%

\section{Introduction\label{sec:Introduction}}

Many image classification models have assisted humans from daily ordinary
tasks like shopping (Google Lens) and entertainment (FaceApp) to important
jobs like healthcare (Calorie Mama) and authentication (BioID).

\begin{figure}
\begin{centering}
\subfloat[]{\begin{centering}
\includegraphics[scale=0.38]{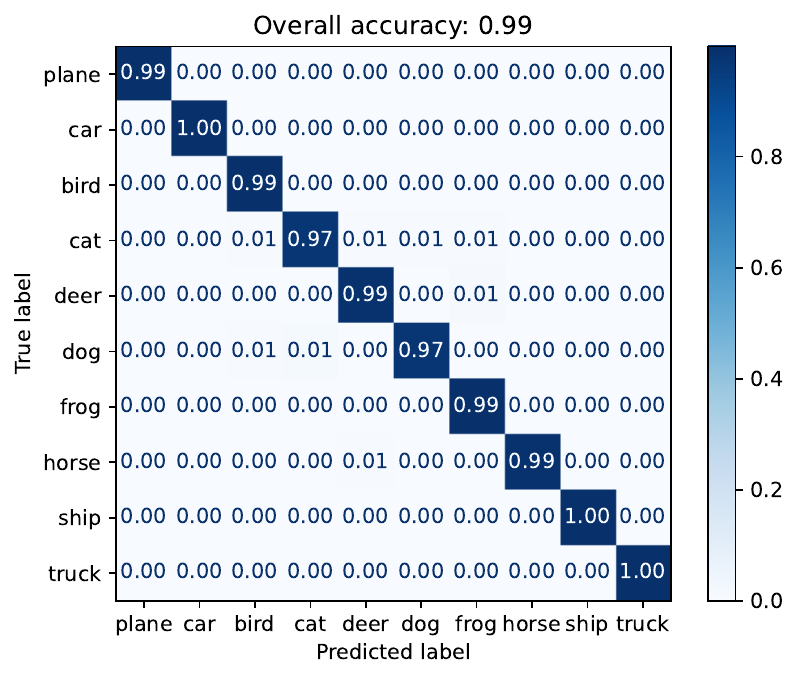}
\par\end{centering}
}\subfloat[]{\begin{centering}
\includegraphics[scale=0.38]{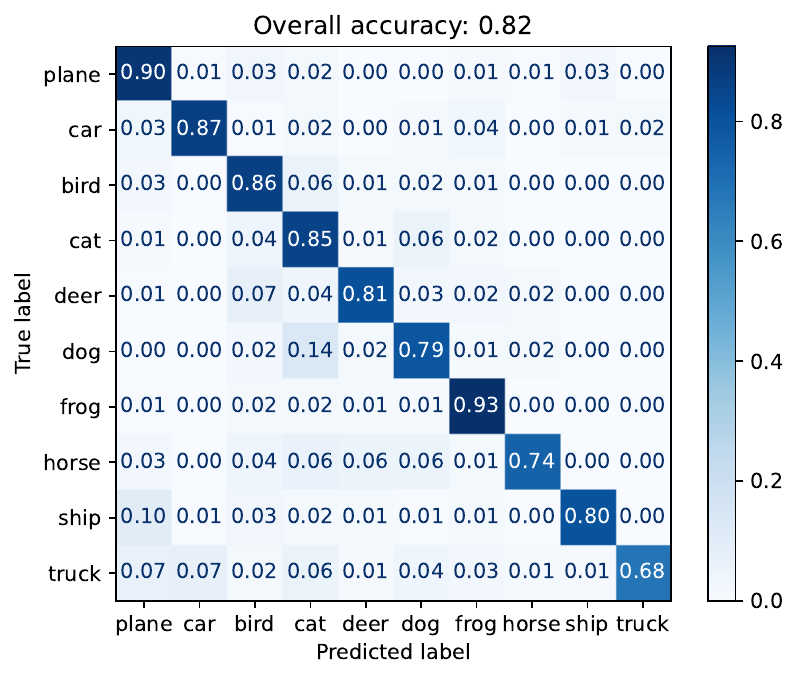}
\par\end{centering}
}\subfloat[]{\begin{centering}
\includegraphics[scale=0.38]{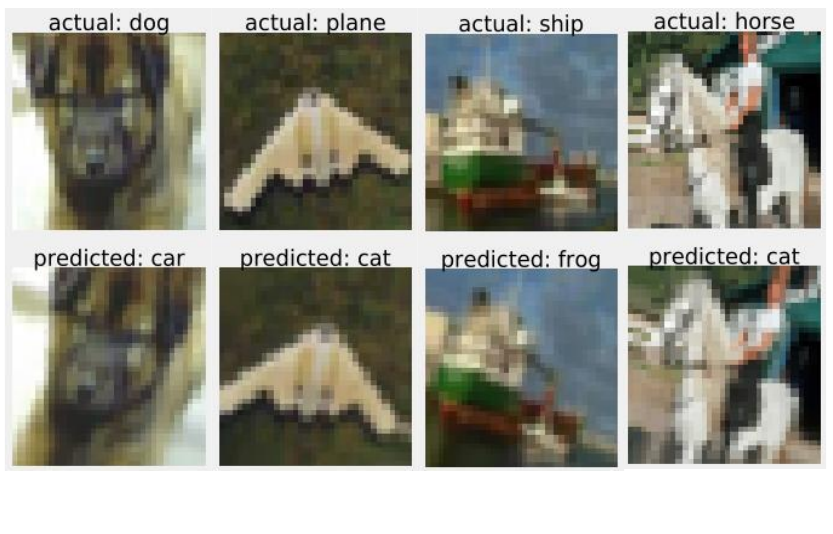}
\par\end{centering}
}
\par\end{centering}
\caption{\label{fig:Model's-predictions-rotate}Overall accuracy and class-wise
accuracy of the ResNet model on original images (a) and distorted
images (b). The overall accuracy dropped 17\% from 0.99 to 0.82 when
the images were rotated $20^{\circ}$. Some misclassified images are
shown in (c).}
\end{figure}

A well-known weakness of \textit{image-classifiers} is that they are
often vulnerable to image distortions i.e. their performance significantly
drops if the input images are distorted \citep{Li2019}. As illustrated
in Figure \ref{fig:Model's-predictions-rotate}, a ResNet model achieved
99\% accuracy on a set of CIFAR-10 images. When the images were slightly
rotated, its accuracy dropped to 82\%. It predicted wrong labels for
$20^{\circ}$ rotated images although these images were easily recognized
by humans. In practice, input images can be distorted in various forms
e.g. rotated images due to an unstable camera, dark images due to
a poor lighting condition, noisy images due to a rainy weather, etc.

For a quality control purpose, we need to evaluate the image-classifier
under different distortion levels to check in which cases it is unreliable/reliable.
As this task is very time- and cost-consuming, it is important to
perform it automatically using a machine learning (ML) approach. In
particular, we ask the question \textquotedblleft \textit{can we predict
the model reliability under an image distortion?}\textquotedblright ,
and form it as a \textit{binary classification} problem. Assume that
we have an \textit{image-classifier} $T$ and a set of labeled images
${\cal D}$ (we call it \textit{verification set}). We define \textit{the
search space of distortion levels} ${\cal C}$ as follows: (1) each
dimension of $\mathcal{C}$ is a\textit{ distortion type} (e.g. rotation,
brightness) and (2) each point $c\in\mathcal{C}$ is a\textit{ distortion
level} (e.g. \{rotation=20$^{\circ}$, brightness=0.5\}), which is
used to modify the images in ${\cal D}$ to create a set of distorted
images ${\cal D}'_{c}$. The model $T$ is called ``reliable'' under
a distortion level $c$ if $T$'s accuracy on ${\cal D}_{c}'$ is
above a stipulated threshold $h$, otherwise ``non-reliable''. Recall
the earlier example, if we choose the threshold $h=95\%$, then the
ResNet model is non-reliable under $20^{\circ}$-rotation as its accuracy
is only 82\%. In other words, the distortion level $20^{\circ}$-rotation
has a label ``non-reliable''. Our goal is to build a \textit{distortion-classifier}
$S$ that receives a distortion level $c\in{\cal C}$ and classifies
it as $0$ (``non-reliable'') or $1$ (``reliable''). For simplicity,
\textit{we treat ``non-reliable'' as negative label whereas ``reliable''
as positive label}.

\begin{figure}
\begin{centering}
\includegraphics[scale=0.5]{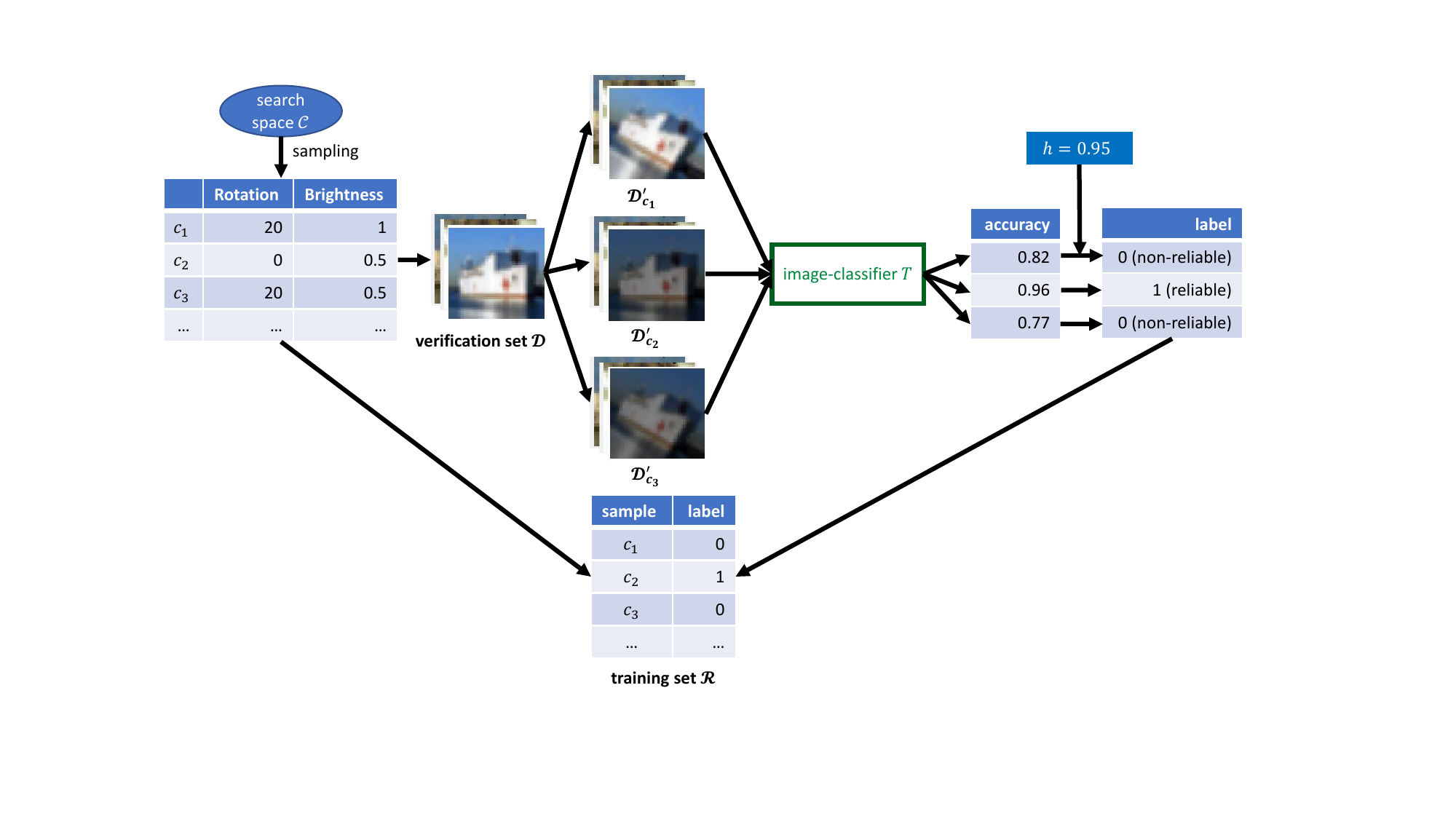}
\par\end{centering}
\caption{\label{fig:Construction-of-R}Construction of the training set ${\cal R}$.
For example, three distortion levels $c_{1}$, $c_{2}$, and $c_{3}$
are randomly sampled from the search space ${\cal C}$ with two dimensions
\{rotation, brightness\}. The label of $c_{1}$ is computed as follows.
First, $c_{1}$ is used to modify images in the verification set ${\cal D}$
to create the set of distorted images ${\cal D}_{c_{1}}'$. Then,
the image-classifier $T$ is evaluated on ${\cal D}_{c_{1}}'$ to
compute its accuracy (0.82). Finally, $c_{1}$ has a label $0$ (i.e.
\textquotedblleft non-reliable\textquotedblright ) as $T$'s accuracy
(0.82) is below the threshold $h=0.95$. The pair $(c_{1},0)$ will
be a sample of the training set ${\cal R}$.}
\end{figure}

The process to train the distortion-classifier $S$ consists of three
steps. (1) \textbf{Construct a training set:} a typical way to create
a training set for $S$ is to randomly sample distortion levels from
the search space ${\cal C}$ and computing their corresponding labels.
Given a $c_{i}\in{\cal C}$, we compute the accuracy $a_{i}$ of the
model $T$ on the set of distorted images ${\cal D}'_{c_{i}}$. If
$a_{i}\geq h$, we assign $c_{i}$ a label ``1'', otherwise a label
``0''. As a result, we obtain a training set ${\cal R}=\{c_{i},\mathbb{I}_{a_{i}\geq h}\}_{i=1}^{I}$,
where $\mathbb{I}_{a_{i}\geq h}$ is an indicator function and $I$
is the sampling budget. We illustrate this procedure to construct
the training set ${\cal R}$ in Figure \ref{fig:Construction-of-R}.
(2) \textbf{Rebalance the training set:} using random distortion levels
often leads to an \textit{imbalanced} training set ${\cal R}$ as
a majority of them fall under negative class, especially when the
threshold $h$ is high or model performance under distortion is generally
poor. Thus, we need to rebalance ${\cal R}$ using an imbalance handling
technique like SMOTE \citep{chawla2002smote}, NearMiss \citep{mani2003knn},
or generative models \citep{sampath2021survey}. (3) \textbf{Train
the distortion-classifier:} we use the rebalanced version of ${\cal R}$
to train a ML predictive model e.g. neural network to classify unseen
distortion levels.

Although current imbalance handling methods can rebalance the training
data, they often do not generate diverse synthetic positive samples
due to a small number of real positive samples in ${\cal R}$. This
leads to a sub-optimal training set for the distortion-classifier
$S$. In this paper, we improve the training set ${\cal R}$ of $S$
with a \textit{Gaussian process (GP) based sampling }technique to
create a training set ${\cal R}$ with a higher fraction of real positive
samples.

\textbf{GP-based sampling:} we consider the mapping from a distortion
level $c$ to the model's accuracy on the set of distorted images
${\cal D}'_{c}$ as a \textit{black-box, expensive function} $f:{\cal C}\rightarrow[0,1]$.
The function $f$ is black-box as we do not know its expression, and
$f$ is expensive as we have to measure the model's accuracy over
all distorted images in ${\cal D}'_{c}$. We approximate $f$ using
a GP \citep{Shahriari2016,Nguyen2020} that is a popular method to
model black-box, expensive functions. We use $f$'s predictive distribution
to design an acquisition function to search for distortion levels
that have a high chance to be a positive sample. We update the GP
with new samples, and repeat the sampling process until the sampling
budget $I$ is depleted. Finally, we obtain a training set ${\cal R}=\{c_{t},\mathbb{I}_{f(c_{t})\geq h}\}_{t=1}^{I}$.

To summarize, we make the following contributions.
\begin{enumerate}
\item We are the first to define the problem of \textit{Prediction of Model
Reliability under Image Distortion}, and propose a distortion-classifier
to predict if the model is reliable under a distortion level.
\item We propose a \textit{GP-based sampling technique} to construct a training
data for the distortion-classifier with an increased fraction of real
positive samples.
\item We extensively evaluate our method on six benchmark image datasets,
and compare it with several strong baselines. We show that it is significantly
better than other methods.
\item The significance of our work lies in providing ability to predict
reliability of \textit{any} image-classifier under a variety of distortions
on \textit{any} image dataset.
\end{enumerate}
The remainder of the paper is organized as follows. In Section \ref{sec:Related-Work},
related works on image distortion, model reliability, and imbalance
classification are reviewed. Our main contributions are presented
in Section \ref{sec:Framework}, where we describe a GP-based sampling
method. Experimental results are discussed in Sections \ref{sec:Experiments}
while conclusions and future works are represented in Section \ref{sec:Conclusion}.

\section{Related Work\label{sec:Related-Work}}

\textbf{Image distortion.} Most deep learning models are sensitive
to image distortion, where a small amount of distortion can severely
reduce their performance. Many methods have been proposed to detect
and correct the distortion in the input images \citep{ahn2017image,Li2019},
which can be categorized into two groups: non-reference and full-reference.
The non-reference methods corrected the distortion without any direct
comparison between the original and distorted images \citep{kang2014convolutional,bosse2016deep}.
Other works developed models that were robust to image distortion,
where most of them fine-tuned the pre-trained models on a pre-defined
set of distorted images \citep{zhou2017classification,dodge2018quality,hossain2019distortion}.
While these methods focused on improving the model quality, which
is useful for the \textit{model development phase}, our work focuses
on predicting the model reliability, which is useful for the\textit{
quality control phase}.

\textbf{Model reliability prediction.} Assessing the reliability of
a ML model is an important step in the quality control process \citep{thung2012empirical}.
Existing works on model reliability focus on defect/bug prediction
\citep{jafarinejad2021nerdbug}, where they classify a model as ``defective''
if its source code has bugs. A typical solution has three main steps
\citep{wang2018software,giray2023use}. First, we collect both ``clean''
and ``defective'' code samples from the model repository to construct
a training set. Second, we rebalance the training set. Finally, we
train a ML predictive model with the rebalanced training set.

Some works target to other reliability aspects of a model such as
relevance and reproducibility \citep{morovati2023bugs}. However,
\textit{there is no work addressing the problem of model reliability
prediction under image distortion}.

\textbf{Imbalance classification.} As the problem of classification
with imbalanced data has been studied for many years, the imbalance
classification has a rich literature. Most existing methods are based
on SMOTE (Synthetic Minority Oversampling Technology) \citep{chawla2002smote},
where the synthetic minority samples are generated by linearly combining
two real minority samples. Several SMOTE variants have been developed
to address SMOTE weaknesses such as outlier and noisy \citep{batista2003balancing,batista2004study,han2005borderline,nguyen2011borderline,sauglam2022novel}.
Other approaches for rebalancing data are under-sampling techniques
\citep{mani2003knn}, ensemble methods \citep{liu2020self-paced-ensemble,liu2020mesa},
and generative models \citep{kingma2019introduction,xu2019modeling,goodfellow2020generative}.

\section{Framework\label{sec:Framework}}

\subsection{Problem statement\label{subsec:Problem-statement}}

Let $T$ be an \textit{image-classifier}, ${\cal D}=\{x_{i},y_{i}\}_{i=1}^{N}$
be a set of labeled images (i.e. \textit{verification set}), and ${\cal E}=\{E_{1},...,E_{d}\}$
be a set of $d$ image distortions e.g. rotation, brightness, etc.
Each $E_{i}$ has a value range $[l_{E_{i}},u_{E_{i}}]$, where $l_{E_{i}}$
and $u_{E_{i}}$ are the lower and upper bounds. We define a compact
subset ${\cal C}$ of $\mathbb{R}^{d}$ as a set of all possible values
for image distortion (i.e. ${\cal C}$ is the search space of all
possible distortion levels).

We consider a mapping function $f:{\cal C}\rightarrow[0,1]$, which
receives a distortion level $c\in{\cal C}$ as input and returns the
accuracy of $T$ on the set of distorted images ${\cal D}_{c}'=\{x_{i}',y_{i}\}_{i=1}^{N}$
as output. Here, each image $x_{i}'\in{\cal D}_{c}'$ is a distorted
version of an original image $x_{i}\in{\cal D}$, caused by the distortion
level $c$. Given a threshold $h\in[0,1]$, $T$ is considered ``reliable''
under $c$ if $f(c)\geq h$, otherwise ``non-reliable''. Without
any loss in generality, \textit{we treat ``non-reliable'' as negative
label (i.e. class $0$) while ``reliable'' as positive label (i.e.
class $1$)}. 

Our goal is to build a \textit{distortion-classifier} $S$ to classify
any distortion level $c\in{\cal C}$ into positive or negative class.

\subsection{Proposed method\label{subsec:Proposed-method}}

The \textit{distortion-classifier} $S$ is trained with three main
steps. First, we create a training set ${\cal R}=\{c_{i},\mathbb{I}_{f(c_{i})\geq h}\}_{i=1}^{I}$,
where $c_{i}$ is randomly sampled from ${\cal C}$ and $I$ is the
sampling budget. However, ${\cal R}$ is often \textit{highly unbalanced},
where the number of negative samples is much more than the number
of positive samples. Second, we rebalance ${\cal R}$ using an imbalance
handling technique such as SMOTE or a generative model. Finally, we
use the rebalanced version of ${\cal R}$ to train $S$ that can be
any ML predictive model e.g. random forest, neural network, etc.

We improve the quality of the training set ${\cal R}$ by proposing
a new sampling approach. Instead of using random sampling method,
we propose a \textit{GP-based sampling }method to sample $c_{i}$
to construct ${\cal R}$.

\subsubsection{GP-based sampling}

Our goal is to sample more positive samples when constructing ${\cal R}$.
To achieve this, we consider the mapping function $f:{\cal C}\rightarrow[0,1]$
as a black-box function and approximate it using a GP. We then use
the GP predictive distribution to guide our sampling process. The
detailed steps are described as follows.
\begin{enumerate}
\item We initialize the training set ${\cal R}_{t}$ with a small set of
randomly sampled distortion levels $[c_{1},...,c_{t}]$ and compute
their function values $\boldsymbol{f}_{1:t}=[f(c_{1}),...,f(c_{t})]$,
where $t$ is a small number i.e. $t\ll I$ (recall that $I$ is the
sampling budget).
\item We use ${\cal R}_{t}=\{c_{i},f(c_{i})\}_{i=1}^{t}$ to learn a GP
to approximate $f$. We assume that $f$ is a smooth function drawn
from a GP, i.e. $f(c)\sim\text{GP}(m(c),k(c,c'))$, where $m(c)$
and $k(c,c')$ are the mean and covariance functions. We compute the
\textit{predictive distribution} for $f(c)$ at any point $c$ as
a Gaussian distribution, with its mean and variance functions:
\begin{align}
\mu_{t}(c) & =\transp{\boldsymbol{k}}K^{-1}\boldsymbol{f}_{1:t}\label{eq:mean_function}\\
\sigma_{t}^{2}(c) & =k(c,c)-\transp{\boldsymbol{k}}K^{-1}\boldsymbol{k}\label{eq:variance_function}
\end{align}
where $\boldsymbol{k}$ is a vector with its $i$-th element defined
as $k(c_{i},c)$ and $K$ is a matrix of size $t\times t$ with its
$(i,j)$-th element defined as $k(c_{i},c_{j})$.
\item We iteratively update the training set ${\cal R}_{t}$ by adding the
new point $\{c_{t+1},f(c_{t+1})\}$ until the sampling budget $I$
is depleted, and at each iteration we also update the GP. Instead
of randomly sampling $c_{t+1}$, we select $c_{t+1}$ by maximizing
the following acquisition function $q(c)$:
\begin{align}
q(c) & =\beta\times\sigma(c)+(\mu(c)-h),\label{eq:acq-function}\\
c_{t+1} & =\argmax{c\in{\cal C}}q(c)\label{eq:suggestion}
\end{align}
where $\mu(c)$ and $\sigma(c)$ are the predictive mean and standard
deviation from Equations (\ref{eq:mean_function}) and (\ref{eq:variance_function}).
The coefficient $\beta=2\times[\log(d\times t\times\pi^{2})-\log(6\times\delta)]$
is computed following \citep{srinivas2012information}, where $d$
is the number of dimensions of the search space ${\cal C}$ and $\delta=0.1$
is a small constant.
\item We construct the training set ${\cal R}=\{c_{t},\mathbb{I}_{f(c_{t})\geq h}\}_{t=1}^{I}$,
where $c_{t}$ is sampled using our acquisition function in Equation
(\ref{eq:suggestion}).
\end{enumerate}
Our sampling strategy achieves two goals: (1) sampling $c$ where
the model\textquoteright s accuracy is higher than the threshold $h$
(i.e. large $(\mu(c)-h)$) and (2) sampling $c$ where the model\textquoteright s
accuracy is highly uncertain (i.e. large $\sigma(c)$). As a result,
we can efficiently find more positive samples. In the experiments,
we show that our GP-based sampling method retrieves a much higher
fraction of positive samples than the random sampling method.

\textbf{Discussion.} We want to highlight that our sampling strategy
is very flexible. If $f(c)\geq h$ is the \textit{minority class}
as in our setting, then we use $(\mu(c)-h)$. If $f(c)<h$ is the
\textit{minority class}, then we can simply change it to $(h-\mu(c))$.

\section{Experiments\label{sec:Experiments}}

\subsection{Experiment settings}

\begin{figure}
\begin{centering}
\includegraphics[scale=0.5]{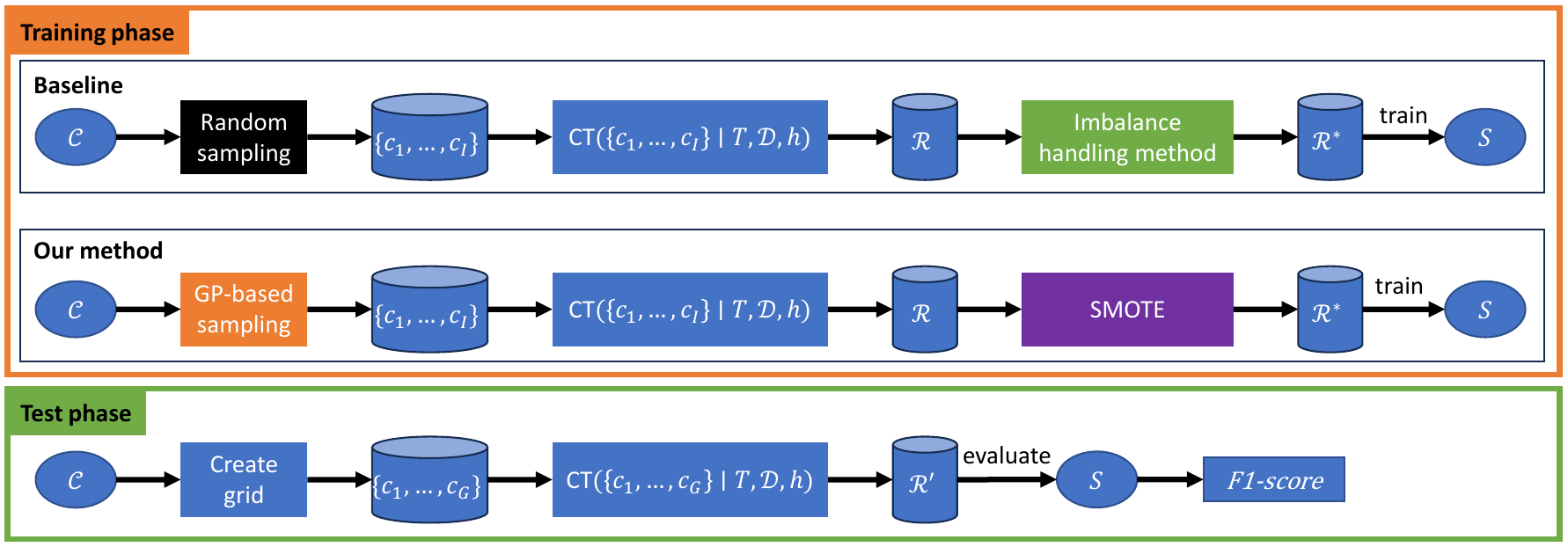}
\par\end{centering}
\caption{\label{fig:Training-test-phases}Steps involved in the training and
test phases of a model reliability prediction task. The module $\text{CT}(\{c_{1},...,c_{I}\}\mid T,{\cal D},h)$
to construct the training set ${\cal R}$ is described in Figure \ref{fig:Construction-of-R}.
There is one difference between our method and the baseline: the GP-based
sampling.}
\end{figure}

We recall steps involved in the training and test phases of our prediction
task for model reliability under image distortion in Figure \ref{fig:Training-test-phases}.
We then provide their implementation details.

\begin{figure}[H]
\begin{centering}
\includegraphics[scale=0.8]{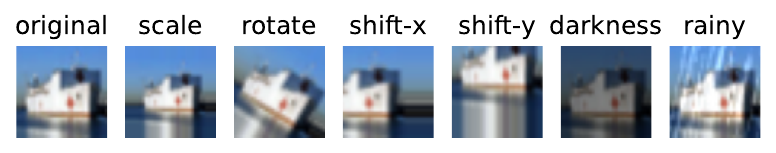}
\par\end{centering}
\caption{\label{fig:Six-distortion-types}Original image plus six distortion
types.}
\end{figure}

\textbf{Search space of distortion levels ${\cal C}$.} We predict
the reliability of image-classifiers against six image distortions
including \textit{geometry distortions} \citep{Gopakumar2018}, \textit{lighting
distortion} \citep{sellahewa2010image}, and \textit{rain distortion}
\citep{patil2022video}. We illustrate six distortion types in Figure
\ref{fig:Six-distortion-types}. The value range of each image distortion
is shown in Table \ref{tab:List-of-transformations}. \textit{Note
that our method is applicable to any distortion types as long as they
can be defined by a range of values}.

\begin{table}[th]
\caption{\label{tab:List-of-transformations}List of distortions along with
their domains.}

\centering{}%
\begin{tabular}{|l|r|l|}
\hline 
\textbf{Distortion} & \textbf{Domain} & \textbf{Description}\tabularnewline
\hline 
\hline 
Scale & $[0.7,1.3]$ & Zoom in/out 0-30\%\tabularnewline
\hline 
Rotation & $[0,90]$ & Rotate $0^{\circ}$ - $90^{\circ}$\tabularnewline
\hline 
Translation-X & $[-0.2,0.2]$ & Shift left/right 0-20\%\tabularnewline
\hline 
Translation-Y & $[-0.2,0.2]$ & Shift up/down 0-20\%\tabularnewline
\hline 
Darkness & $[0.7,1.3]$ & Darken/brighten 0-30\%\tabularnewline
\hline 
Rain & $[0,1]$ & $0$: no rain, $1$: a lot of rain\tabularnewline
\hline 
\end{tabular}
\end{table}

\textbf{Sampling method.} While the baseline uses a random sampling,
our method uses the GP-based sampling. After the sampling process,
we obtain a set of distortion levels $\{c_{1},...,c_{I}\}$, where
$I=600$ is the sampling budget.

\textbf{Construction of training set ${\cal R}$.} Given distortion
levels $\{c_{1},...,c_{I}\}$, the module $\text{CT}(\{c_{1},...,c_{I}\}\mid T,{\cal D},h)$
assigns label $0$ or $1$ for each $c_{i}$ (see Figure \ref{fig:Construction-of-R}).
It requires image-classifier $T$, verification set ${\cal D}$, and
reliability threshold $h$. 

For image-classifiers $T$, we use five pre-trained models from \citep{Nguyen2022}
for image datasets MNIST, Fashion, CIFAR-10, CIFAR-100, and Tiny-ImageNet.
They achieved similar accuracy as those reported in \citep{Tian2020,Wang2020,Nguyen2021,bhat2021distill}.
We also use the pre-trained ResNet50 model from the Keras library\footnote{https://keras.io/api/applications/resnet/\#resnet50-function}
for ImageNette\footnote{https://www.tensorflow.org/datasets/catalog/imagenette}.
As pointed out by \citep{hossain2019distortion,Li2019}, we expect
that these image-classifiers will reduce their performance when evaluated
on distorted images.

For each image dataset, we use 10\% of its data samples to be the
verification set ${\cal D}$. The size of ${\cal D}$, the accuracy
of $T$ on ${\cal D}$, and the reliability threshold $h$ are shown
in Table \ref{tab:Verification-set-and-Reliability-threshold}.

\begin{table}[th]
\caption{\label{tab:Verification-set-and-Reliability-threshold}Size of verification
set ${\cal D}$, accuracy of $T$ on ${\cal D}$, and reliability
threshold $h$ used in our experiments.}

\centering{}%
\begin{tabular}{|l|r|r|r|}
\hline 
 & \textbf{$|{\cal D}|$} & \textbf{Accuracy of $T$} & \textbf{$h$}\tabularnewline
\hline 
\hline 
MNIST & 6,000 & 0.9967 & 0.90\tabularnewline
\hline 
Fashion & 6,000 & 0.9908 & 0.75\tabularnewline
\hline 
CIFAR-10 & 5,000 & 0.9902 & 0.85\tabularnewline
\hline 
CIFAR-100 & 5,000 & 0.9340 & 0.65\tabularnewline
\hline 
Tiny-ImageNet & 10,000 & 0.6275 & 0.45\tabularnewline
\hline 
ImageNette & 1,000 & 0.8290 & 0.70\tabularnewline
\hline 
\end{tabular}
\end{table}

\textbf{Imbalance handling method.} As the training set ${\cal R}$
is highly imbalanced, we rebalance it before training the distortion-classifier
$S$. We use \textit{SMOTE} \citep{chawla2002smote} and compare it
with SOTA imbalance handling methods, including \textit{Cost-sensitive
learning} \citep{thai2010cost}, under-sampling method \textit{NearMiss}
\citep{mani2003knn}, over-sampling methods \textit{AdaSyn} \citep{he2008adasyn},
ensemble methods \textit{SPE} \citep{liu2020self-paced-ensemble}
and \textit{MESA} \citep{liu2020mesa}, and generative models \textit{GAN}
\citep{goodfellow2020generative}, \textit{VAE} \citep{kingma2019introduction},
\textit{CTGAN}, and \textit{TVAE} \citep{xu2019modeling}.

As our method is SMOTE, we also compare it with SMOTE variants, including
\textit{SMOTE-Borderline} \citep{han2005borderline}, \textit{SMOTE-SVM}
\citep{nguyen2011borderline}, \textit{SMOTE-ENN} \citep{batista2004study},
\textit{SMOTE-TOMEK} \citep{batista2003balancing}, and \textit{SMOTE-WB}
\citep{sauglam2022novel}. To be fair, we use the source codes released
by the authors or implemented in well-known public libraries. The
details are provided in \ref{sec:Implementation-baselines}.

\textbf{Distortion-classifier $S$.} We train five popular ML predictive
models with the rebalanced training set ${\cal R}^{*}$, including
\textit{decision tree}, \textit{random forest}, \textit{logistic regression},
\textit{support vector machine}, and \textit{neural network}. Each
of them is a distortion-classifier. In the test phase, we report the
averaged result of five distortion-classifiers.

\textbf{Construction of test set ${\cal R}'$.} To evaluate the performance
of distortion-classifiers, we need to construct a test set. We create
a grid of distortion levels $\{c_{1},...,c_{G}\}$ in ${\cal C}$.
For each dimension, we use five points, resulting in 4,096 grid points
in total. For each test point $c$, we determine its label using the
procedure in Figure \ref{fig:Construction-of-R}. At the end, there
are 4,096 test distortion levels along with their labels. We report
the numbers of positive and negative test points in \ref{sec:Test-set}.

\textbf{Evaluation metric.} We evaluate each distortion-classifier
on the test set ${\cal R}'$ and compute the F1-score. As each imbalance
handling method is combined with five ML predictive models to form
five distortion-classifiers, we report the averaged F1-score. A higher
F1-score means a better prediction.

We repeat each method three times with random seeds, and report the
averaged F1-score. As the standard deviations are small (< 0.06),
we do not report them to save space.

\subsection{Comparison of sampling methods}

\begin{figure}[th]
\begin{centering}
\includegraphics[scale=0.55]{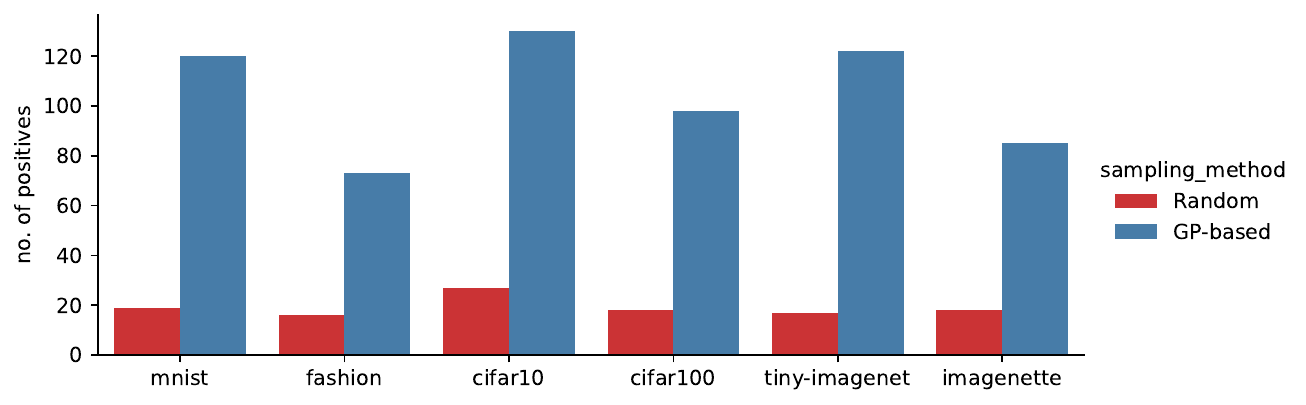}
\par\end{centering}
\caption{\label{fig:Sampling-method}Number of real positive samples sampled
from ${\cal C}$ by the random sampling and our GP-based sampling.}
\end{figure}

We compare our GP-based sampling with the random sampling. From Figure
\ref{fig:Sampling-method}, our GP-based sampling obtains many more
positive points than the random sampling. For example, on CIFAR-10,
among 600 sampled points, the random sampling obtains only 27 positive
samples to construct the training set ${\cal R}$. In contrast, our
GP-based sampling retrieves 130 positive samples to construct a more
balanced ${\cal R}$.

\subsection{Comparison of imbalance handling methods}

We compare our imbalance handling method GP-based sampling + SMOTE
(we call it \textbf{GPS-SMOTE}) with current state-of-the-art imbalance
handling methods.

Table \ref{tab:Imbalance-handling} shows that our GPS-SMOTE is the
best method and significantly outperforms other methods. Its improvements
are around 2\% on MNIST, 8\% on Fashion, 2\% on CIFAR-10, 6\% on CIFAR-100,
2\% on Tiny-ImageNet, and 8\% on ImageNette.

\begin{table}
\caption{\label{tab:Imbalance-handling}F1-scores of our method GPS-SMOTE and
other imbalance handling methods. Datasets include \textbf{M:} MNIST,
\textbf{F:} Fashion, \textbf{C10:} CIFAR-10, \textbf{C100:} CIFAR-100,
\textbf{T-IN:} Tiny-ImageNet, and \textbf{IN:} ImageNette.}

\centering{}%
\begin{tabular}{|l|l|l|r|r|r|r|r|r|}
\hline 
 & \textbf{Sampling} & \textbf{Imbalance} & \textbf{M} & \textbf{F} & \textbf{C10} & \textbf{C100} & \textbf{T-IN} & \textbf{IN}\tabularnewline
\hline 
Standard & Random & None & 0.3657 & 0.2105 & 0.6507 & 0.4130 & 0.3587 & 0.6561\tabularnewline
\hline 
Cost-sensitive & Random & Re-weight & 0.5478 & 0.3940 & 0.6938 & 0.5593 & 0.5256 & 0.6677\tabularnewline
\hline 
\multirow{2}{*}{Under-sampling} & Random & RandomUnder & 0.2553 & 0.1467 & 0.5531 & 0.2824 & 0.2870 & 0.3989\tabularnewline
\cline{2-9}
 & Random & NearMiss & 0.3358 & 0.1514 & 0.6588 & 0.4610 & 0.3443 & 0.6764\tabularnewline
\hline 
\multirow{3}{*}{Over-sampling} & Random & RandomOver & 0.6194 & 0.4554 & 0.7230 & 0.5939 & 0.5797 & 0.7063\tabularnewline
\cline{2-9}
 & Random & SMOTE & 0.6157 & 0.4379 & 0.7310 & 0.5933 & 0.5658 & 0.7100\tabularnewline
\cline{2-9}
 & Random & Adasyn & 0.6090 & 0.4370 & 0.7306 & 0.5955 & 0.5663 & 0.7065\tabularnewline
\hline 
\multirow{2}{*}{Ensemble} & Random & SPE & 0.5237 & 0.2984 & 0.7269 & 0.5113 & 0.4816 & 0.6808\tabularnewline
\cline{2-9}
 & Random & MESA & 0.4337 & 0.2120 & 0.6402 & 0.4394 & 0.3739 & 0.5802\tabularnewline
\hline 
\multirow{4}{*}{Deep learning} & Random & GAN & 0.3202 & 0.2186 & 0.5157 & 0.3358 & 0.3551 & 0.4739\tabularnewline
\cline{2-9}
 & Random & VAE & 0.3831 & 0.2264 & 0.6635 & 0.4123 & 0.3821 & 0.6638\tabularnewline
\cline{2-9}
 & Random & CTGAN & 0.2958 & 0.1966 & 0.4124 & 0.2968 & 0.3144 & 0.3904\tabularnewline
\cline{2-9}
 & Random & TVAE & 0.3364 & 0.2124 & 0.5319 & 0.3365 & 0.3249 & 0.4673\tabularnewline
\hline 
\multirow{1}{*}{\textbf{GPS-SMOTE (Ours)}} & GP-based & SMOTE & \textbf{0.6361} & \textbf{0.5327} & \textbf{0.7562} & \textbf{0.6525} & \textbf{0.5952} & \textbf{0.7988}\tabularnewline
\hline 
\end{tabular}
\end{table}

Among imbalance handling baselines, SMOTE often achieves the best
results. When SMOTE is combined with our GP-based sampling, its performance
is improved significantly. This shows that our GP-based sampling is
better than the random sampling.

In general, imbalance handling methods often improve the performance
of the distortion-classifier, compared to the standard distortion-classifier.
Over-sampling methods are always better than under-sampling methods.
Deep learning methods based on generative models do not show any real
benefit.

\textbf{Comparison with SMOTE variants.} We also compare our GPS-SMOTE
with imbalance handling methods based on SMOTE in Table \ref{tab:SMOTE-variants}.
Our method is the best method while other SMOTE-based methods perform
similarly.

\begin{table}
\caption{\label{tab:SMOTE-variants}F1-scores of our method GPS-SMOTE and SMOTE
variants.}

\centering{}%
\begin{tabular}{|l|r|r|r|r|r|r|}
\hline 
 & \textbf{MNIST} & \textbf{Fashion} & \textbf{CIFAR-10} & \textbf{CIFAR-100} & \textbf{Tiny-ImageNet} & \textbf{ImageNette}\tabularnewline
\hline 
\hline 
SMOTE & 0.6157 & 0.4379 & 0.7310 & 0.5933 & 0.5658 & 0.7100\tabularnewline
\hline 
SMOTE-Borderline & 0.6118 & 0.4313 & 0.7246 & 0.5924 & 0.5673 & 0.7057\tabularnewline
\hline 
SMOTE-SVM & 0.6120 & 0.4231 & 0.7335 & 0.6020 & 0.5711 & 0.7187\tabularnewline
\hline 
SMOTE-ENN & 0.6046 & 0.3931 & 0.6752 & 0.5504 & 0.5213 & 0.6744\tabularnewline
\hline 
SMOTE-TOMEK & 0.6155 & 0.4381 & 0.7312 & 0.5931 & 0.5658 & 0.7096\tabularnewline
\hline 
SMOTE-WB & 0.6210 & 0.4511 & 0.7276 & 0.5859 & 0.5659 & 0.7079\tabularnewline
\hline 
\textbf{GPS-SMOTE (Ours)} & \textbf{0.6361} & \textbf{0.5327} & \textbf{0.7562} & \textbf{0.6525} & \textbf{0.5952} & \textbf{0.7988}\tabularnewline
\hline 
\end{tabular}
\end{table}

\subsection{Ablation study}

We conduct further experiments on CIFAR-10 to analyze our method under
different settings.

\textbf{Sampling budget $I$.} We investigate the effect of the number
of sampling queries (i.e. the size of the sampling budget $I$) on
the performance of our method.

\begin{figure}[H]
\begin{centering}
\includegraphics[scale=0.55]{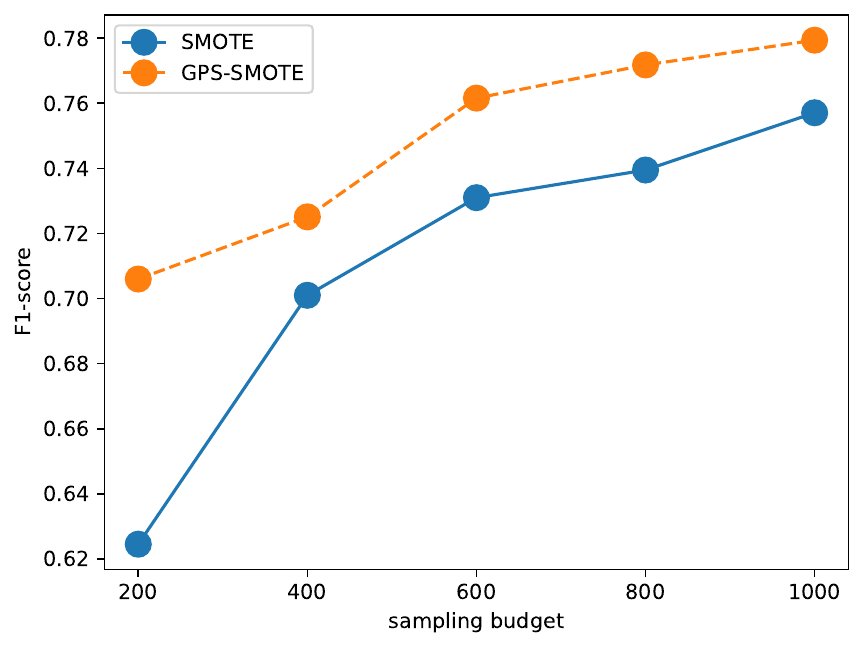}\vspace{-0.2cm}
\par\end{centering}
\caption{\label{fig:Sampling-budget}Our F1-score vs. the sampling budget $I$.}
\end{figure}

Figure \ref{fig:Sampling-budget} shows that both methods improve
as the number of sampling queries increase as expected. More queries
result in more training data and more chance to get positive samples.
However, our GPS-SMOTE is always better than SMOTE by a large margin.

\textbf{Reliability threshold $h$.} We investigate how our performance
is changed with different reliability thresholds $h$.

\begin{figure}[H]
\begin{centering}
\includegraphics[scale=0.55]{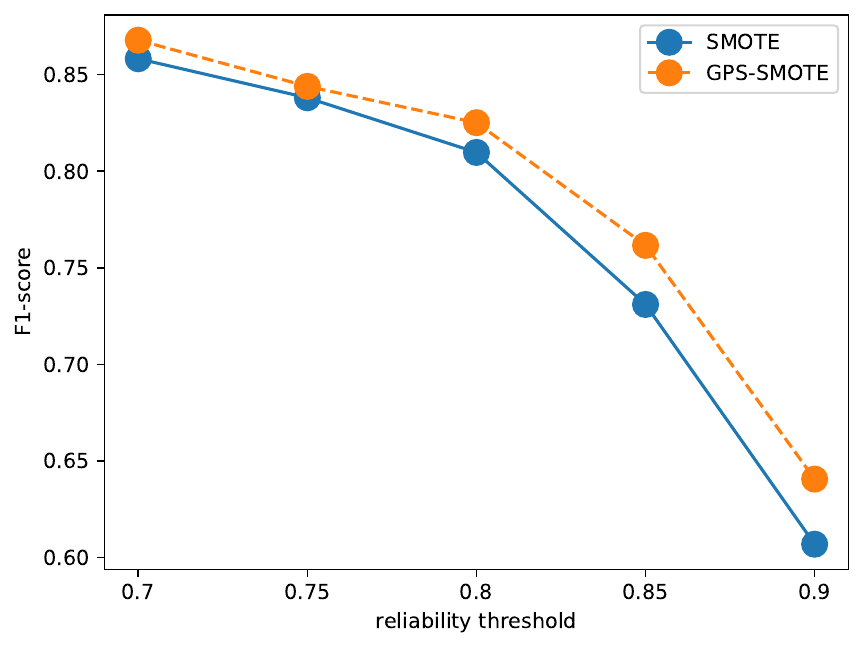}\vspace{-0.2cm}
\par\end{centering}
\caption{\label{fig:Reliability-threshold}Our F1-score vs. the reliability
threshold $h$.}
\end{figure}

Figure \ref{fig:Reliability-threshold} shows that both methods reduce
their F1-scores when the reliability threshold $h$ becomes larger
as the image-classifier $T$ is reliable under fewer distortion levels
(i.e. fewer positive samples). This leads to a \textit{very highly
imbalanced} training set ${\cal R}$.

\section{Conclusion\label{sec:Conclusion}}

Predicting model reliability is an important task in the quality control
process. In this paper, we solve this task in the context of image
distortion i.e. we predict if an image-classifier is unreliable/reliable
under a distortion level. We form this task as a binary classification
process with three main steps: (1) construct a training set, (2) rebalance
the training set, and (3) train a distortion-classifier. As the training
set is highly imbalanced, we propose a method to handle the class-imbalance:
a GP-based sampling. Namely, we approximate the black-box function
mapping from a distortion level to the model's accuracy on distorted
images using GP. We then leverage the GP's mean and variance to form
our sampling process. We demonstrate the benefits of our method on
six image datasets, where it greatly outperforms other baselines.

\textbf{Future works.} We will investigate and verify our method for
other ML and deep learning models including large language models
\citep{chang2023survey}, generative models \citep{goodfellow2020generative},
and tabular models \citep{gorishniy2021revisiting,nguyen2024tabular}.

\bibliographystyle{plain}
\bibliography{reference}

\begin{thebibliography}{10}

\bibitem{ahn2017image}
Namhyuk Ahn, Byungkon Kang, and Kyung-Ah Sohn.
\newblock Image distortion detection using convolutional neural network.
\newblock In {\em IEEE Asian Conference on Pattern Recognition (ACPR)}, pages
  220--225, 2017.

\bibitem{batista2003balancing}
Gustavo Batista, Ana Bazzan, Maria~Carolina Monard, et~al.
\newblock Balancing training data for automated annotation of keywords: a case
  study.
\newblock {\em WoB}, 3:10--8, 2003.

\bibitem{batista2004study}
Gustavo Batista, Ronaldo Prati, and Maria~Carolina Monard.
\newblock A study of the behavior of several methods for balancing machine
  learning training data.
\newblock {\em ACM SIGKDD Explorations Newsletter}, 6(1):20--29, 2004.

\bibitem{bhat2021distill}
Prashant Bhat, Elahe Arani, and Bahram Zonooz.
\newblock Distill on the go: Online knowledge distillation in self-supervised
  learning.
\newblock In {\em CVPR Workshop}, pages 2678--2687, 2021.

\bibitem{bosse2016deep}
Sebastian Bosse, Dominique Maniry, Thomas Wiegand, and Wojciech Samek.
\newblock A deep neural network for image quality assessment.
\newblock In {\em IEEE International Conference on Image Processing (ICIP)},
  pages 3773--3777, 2016.

\bibitem{chang2023survey}
Yupeng Chang, Xu~Wang, Jindong Wang, Yuan Wu, Linyi Yang, Kaijie Zhu, Hao Chen,
  Xiaoyuan Yi, Cunxiang Wang, Yidong Wang, et~al.
\newblock A survey on evaluation of large language models.
\newblock {\em ACM Transactions on Intelligent Systems and Technology}, 2023.

\bibitem{chawla2002smote}
Nitesh Chawla, Kevin Bowyer, Lawrence Hall, and Philip Kegelmeyer.
\newblock {SMOTE: synthetic minority over-sampling technique}.
\newblock {\em Journal of Artificial Intelligence Research}, 16:321--357, 2002.

\bibitem{dodge2018quality}
Samuel Dodge and Lina Karam.
\newblock Quality robust mixtures of deep neural networks.
\newblock {\em IEEE Transactions on Image Processing}, 27(11):5553--5562, 2018.

\bibitem{giray2023use}
G{\"o}rkem Giray, Kwabena~Ebo Bennin, {\"O}mer K{\"o}ksal, {\"O}nder Babur, and
  Bedir Tekinerdogan.
\newblock On the use of deep learning in software defect prediction.
\newblock {\em Journal of Systems and Software}, 195:111537, 2023.

\bibitem{goodfellow2020generative}
Ian Goodfellow, Jean Pouget-Abadie, Mehdi Mirza, Bing Xu, David Warde-Farley,
  Sherjil Ozair, Aaron Courville, and Yoshua Bengio.
\newblock Generative adversarial networks.
\newblock {\em Communications of the ACM}, 63(11):139--144, 2020.

\bibitem{Gopakumar2018}
Shivapratap Gopakumar, Sunil Gupta, Santu Rana, Vu~Nguyen, and Svetha
  Venkatesh.
\newblock Algorithmic assurance: An active approach to algorithmic testing
  using bayesian optimisation.
\newblock In {\em NIPS}, pages 5466--5474, 2018.

\bibitem{gorishniy2021revisiting}
Yury Gorishniy, Ivan Rubachev, Valentin Khrulkov, and Artem Babenko.
\newblock Revisiting deep learning models for tabular data.
\newblock In {\em NeurIPS}, volume~34, pages 18932--18943, 2021.

\bibitem{han2005borderline}
Hui Han, Wen-Yuan Wang, and Bing-Huan Mao.
\newblock {Borderline-SMOTE: a new over-sampling method in imbalanced data sets
  learning}.
\newblock In {\em International Conference on Intelligent Computing}, pages
  878--887, 2005.

\bibitem{he2008adasyn}
Haibo He, Yang Bai, Edwardo Garcia, and Shutao Li.
\newblock {ADASYN: Adaptive synthetic sampling approach for imbalanced
  learning}.
\newblock In {\em IJCNN}, pages 1322--1328, 2008.

\bibitem{hossain2019distortion}
Md~Tahmid Hossain, Shyh~Wei Teng, Dengsheng Zhang, Suryani Lim, and Guojun Lu.
\newblock Distortion robust image classification using deep convolutional
  neural network with discrete cosine transform.
\newblock In {\em IEEE International Conference on Image Processing (ICIP)},
  pages 659--663, 2019.

\bibitem{jafarinejad2021nerdbug}
Foad Jafarinejad, Krishna Narasimhan, and Mira Mezini.
\newblock {NerdBug: automated bug detection in neural networks}.
\newblock In {\em International Workshop on AI and Software Testing/Analysis},
  pages 13--16, 2021.

\bibitem{kang2014convolutional}
Le~Kang, Peng Ye, Yi~Li, and David Doermann.
\newblock Convolutional neural networks for no-reference image quality
  assessment.
\newblock In {\em CVPR}, pages 1733--1740, 2014.

\bibitem{kingma2019introduction}
Diederik Kingma, Max Welling, et~al.
\newblock An introduction to variational autoencoders.
\newblock {\em Foundations and Trends in Machine Learning}, 12(4):307--392,
  2019.

\bibitem{Li2019}
Xiaoyu Li, Bo~Zhang, Pedro Sander, and Jing Liao.
\newblock Blind geometric distortion correction on images through deep
  learning.
\newblock In {\em CVPR}, pages 4855--4864, 2019.

\bibitem{liu2020self-paced-ensemble}
Zhining Liu, Wei Cao, Zhifeng Gao, Jiang Bian, Hechang Chen, Yi~Chang, and
  Tie-Yan Liu.
\newblock Self-paced ensemble for highly imbalanced massive data
  classification.
\newblock In {\em ICDE}, pages 841--852, 2020.

\bibitem{liu2020mesa}
Zhining Liu, Pengfei Wei, Jing Jiang, Wei Cao, Jiang Bian, and Yi~Chang.
\newblock {MESA: Boost Ensemble Imbalanced Learning with MEta-SAmpler}.
\newblock In {\em NeurIPS}, volume~33, pages 14463--14474, 2020.

\bibitem{mani2003knn}
Inderjeet Mani and Jianping Zhang.
\newblock {kNN approach to unbalanced data distributions: a case study
  involving information extraction}.
\newblock In {\em ICML Workshop}, volume 126, pages 1--7, 2003.

\bibitem{morovati2023bugs}
Mohammad~Mehdi Morovati, Amin Nikanjam, Foutse Khomh, and Zhen~Ming Jiang.
\newblock Bugs in machine learning-based systems: a faultload benchmark.
\newblock {\em Empirical Software Engineering}, 28(3):62, 2023.

\bibitem{nguyen2024tabular}
Dang Nguyen, Sunil Gupta, Kien Do, Thin Nguyen, and Svetha Venkatesh.
\newblock Generating realistic tabular data with large language models.
\newblock In {\em ICDM}, 2024.

\bibitem{Nguyen2022}
Dang Nguyen, Sunil Gupta, Kien Do, and Svetha Venkatesh.
\newblock Black-box few-shot knowledge distillation.
\newblock In {\em ECCV}, 2022.

\bibitem{Nguyen2021}
Dang Nguyen, Sunil Gupta, Trong Nguyen, Santu Rana, Phuoc Nguyen, Truyen Tran,
  Ky~Le, Shannon Ryan, and Svetha Venkatesh.
\newblock Knowledge distillation with distribution mismatch.
\newblock In {\em ECML-PKDD}, pages 250--265, 2021.

\bibitem{Nguyen2020}
Dang Nguyen, Sunil Gupta, Santu Rana, Alistair Shilton, and Svetha Venkatesh.
\newblock Bayesian optimization for categorical and category-specific
  continuous inputs.
\newblock In {\em AAAI}, volume~34, pages 5256--5263, 2020.

\bibitem{nguyen2011borderline}
Hien Nguyen, Eric Cooper, and Katsuari Kamei.
\newblock Borderline over-sampling for imbalanced data classification.
\newblock {\em International Journal of Knowledge Engineering and Soft Data
  Paradigms}, 3(1):4--21, 2011.

\bibitem{patil2022video}
Prashant Patil, Sunil Gupta, Santu Rana, and Svetha Venkatesh.
\newblock Video restoration framework and its meta-adaptations to data-poor
  conditions.
\newblock In {\em ECCV}, pages 143--160, 2022.

\bibitem{sauglam2022novel}
Fatih Sa{\u{g}}lam and Mehmet~Ali Cengiz.
\newblock {A novel SMOTE-based resampling technique trough noise detection and
  the boosting procedure}.
\newblock {\em Expert Systems with Applications}, 200:117023, 2022.

\bibitem{sampath2021survey}
Vignesh Sampath, I{\~n}aki Maurtua, Juan~Jose Aguilar~Martin, and Aitor
  Gutierrez.
\newblock A survey on generative adversarial networks for imbalance problems in
  computer vision tasks.
\newblock {\em Journal of Big Data}, 8:1--59, 2021.

\bibitem{sellahewa2010image}
Harin Sellahewa and Sabah Jassim.
\newblock Image-quality-based adaptive face recognition.
\newblock {\em IEEE Transactions on Instrumentation and measurement},
  59(4):805--813, 2010.

\bibitem{Shahriari2016}
Bobak Shahriari, Kevin Swersky, Ziyu Wang, Ryan Adams, and Nando Freitas.
\newblock Taking the human out of the loop: A review of bayesian optimization.
\newblock {\em Proceedings of the IEEE}, 104(1):148--175, 2016.

\bibitem{srinivas2012information}
Niranjan Srinivas, Andreas Krause, Sham Kakade, and Matthias Seeger.
\newblock Information-theoretic regret bounds for gaussian process optimization
  in the bandit setting.
\newblock {\em IEEE Transactions on Information Theory}, 58(5):3250--3265,
  2012.

\bibitem{thai2010cost}
Nguyen Thai-Nghe, Zeno Gantner, and Lars Schmidt-Thieme.
\newblock Cost-sensitive learning methods for imbalanced data.
\newblock In {\em IJCNN}, pages 1--8, 2010.

\bibitem{thung2012empirical}
Ferdian Thung, Shaowei Wang, David Lo, and Lingxiao Jiang.
\newblock An empirical study of bugs in machine learning systems.
\newblock In {\em International Symposium on Software Reliability Engineering},
  pages 271--280, 2012.

\bibitem{Tian2020}
Yonglong Tian, Dilip Krishnan, and Phillip Isola.
\newblock Contrastive representation distillation.
\newblock In {\em ICLR}, 2020.

\bibitem{Wang2020}
Dongdong Wang, Yandong Li, Liqiang Wang, and Boqing Gong.
\newblock {Neural Networks Are More Productive Teachers Than Human Raters:
  Active Mixup for Data-Efficient Knowledge Distillation from a Blackbox
  Model}.
\newblock In {\em CVPR}, pages 1498--1507, 2020.

\bibitem{wang2018software}
Jinyong Wang and Ce~Zhang.
\newblock {Software reliability prediction using a deep learning model based on
  the RNN encoder--decoder}.
\newblock {\em Reliability Engineering \& System Safety}, 170:73--82, 2018.

\bibitem{xu2019modeling}
Lei Xu, Maria Skoularidou, Alfredo Cuesta-Infante, and Kalyan Veeramachaneni.
\newblock {Modeling tabular data using Conditional GAN}.
\newblock In {\em NeurIPS}, volume~32, pages 7335--7345, 2019.

\bibitem{zhou2017classification}
Yiren Zhou, Sibo Song, and Ngai-Man Cheung.
\newblock On classification of distorted images with deep convolutional neural
  networks.
\newblock In {\em IEEE International Conference on Acoustics, Speech and Signal
  Processing (ICASSP)}, pages 1213--1217, 2017.

\end{thebibliography}
\newpage{}

\appendix

\section{Test set ${\cal R}'$\label{sec:Test-set}}

Table \ref{tab:Test-sets-MAID} reports the numbers of negative and
positive samples in the test set ${\cal R}'$, which is used to evaluate
the performance of distortion-classifiers (see Figure \ref{fig:Training-test-phases}).

\begin{table}[H]
\caption{\label{tab:Test-sets-MAID}Test sets ${\cal R}'$ to evaluate distortion-classifiers
in the test phase.}

\centering{}%
\begin{tabular}{|l|r|r|}
\hline 
\textbf{Dataset} & \textbf{\#negative} & \textbf{\#positive}\tabularnewline
\hline 
\hline 
MNIST & 3,957 & 139\tabularnewline
\hline 
Fashion & 4,017 & 79\tabularnewline
\hline 
CIFAR-10 & 3,884 & 212\tabularnewline
\hline 
CIFAR-100 & 3,977 & 119\tabularnewline
\hline 
Tiny-ImageNet & 3,991 & 105\tabularnewline
\hline 
ImageNette & 3,940 & 156\tabularnewline
\hline 
\end{tabular}
\end{table}

\section{Implementation of baselines\label{sec:Implementation-baselines}}

To be fair, when comparing with other methods, we use their source
code released by the authors or their implementation in well-known
public libraries. Table \ref{tab:Implementation-link} shows the link
to the implementation of each baseline.

\begin{table}
\caption{\label{tab:Implementation-link}Method and its implementation link.}

\centering{}%
\begin{tabular}{|l|l|}
\hline 
\textbf{Method} & \textbf{Implementation link}\tabularnewline
\hline 
\hline 
Re-weight & https://scikit-learn.org/stable/\tabularnewline
\hline 
RandomUnder & \multirow{9}{*}{https://imbalanced-learn.org/stable/}\tabularnewline
\cline{1-1}
NearMiss & \tabularnewline
\cline{1-1}
RandomOver & \tabularnewline
\cline{1-1}
SMOTE & \tabularnewline
\cline{1-1}
SMOTE-Borderline & \tabularnewline
\cline{1-1}
SMOTE-SVM & \tabularnewline
\cline{1-1}
SMOTE-ENN & \tabularnewline
\cline{1-1}
SMOTE-TOMEK & \tabularnewline
\cline{1-1}
Adasyn & \tabularnewline
\hline 
SMOTE-WB & https://github.com/analyticalmindsltd/smote\_variants\tabularnewline
\hline 
SPE & https://github.com/ZhiningLiu1998/imbalanced-ensemble\tabularnewline
\hline 
MESA & https://github.com/ZhiningLiu1998/mesa\tabularnewline
\hline 
GAN & https://github.com/dialnd/imbalanced-algorithms\tabularnewline
\hline 
VAE & https://github.com/dialnd/imbalanced-algorithms\tabularnewline
\hline 
CTGAN & https://github.com/sdv-dev/CTGAN\tabularnewline
\hline 
TVAE & https://github.com/sdv-dev/CTGAN\tabularnewline
\hline 
\end{tabular}
\end{table}

\end{document}